\documentclass[sigconf]{acmart}

\usepackage[utf8]{inputenc}
\usepackage{subfig}
\usepackage{booktabs}
\usepackage{pbox}

\AtBeginDocument{%
  \providecommand\BibTeX{{%
    \normalfont B\kern-0.5em{\scshape i\kern-0.25em b}\kern-0.8em\TeX}}}

\setcopyright{acmcopyright}
\begin{document}

\title[How Does BERT Answer Questions?]{How Does BERT Answer Questions?\\A Layer-Wise Analysis of Transformer Representations}

\author{Betty van Aken}
\authornote{Both authors contributed equally to this research.}
\email{bvanaken@beuth-hochschule.de}
\affiliation{%
  \institution{Beuth University of Applied Sciences Berlin}
  \streetaddress{Luxembourg street 10}
  \postcode{13353}
}
\author{Benjamin Winter}
\authornotemark[1]
\email{Benjamin.Winter@beuth-hochschule.de}
\affiliation{%
  \institution{Beuth University of Applied Sciences Berlin}
  \streetaddress{Luxembourg street 10}
  \postcode{13353}
}
\author{Alexander Löser}
\email{aloeser@beuth-hochschule.de}
\affiliation{%
  \institution{Beuth University of Applied Sciences Berlin}
  \streetaddress{Luxembourg street 10}
  \postcode{13353}
}
\author{Felix A. Gers}
\email{gers@beuth-hochschule.de}
\affiliation{%
  \institution{Beuth University of Applied Sciences Berlin}
  \streetaddress{Luxembourg street 10}
  \postcode{13353}
}

\begin{abstract}
Bidirectional Encoder Representations from Transformers (BERT) reach state-of-the-art results in a variety of Natural Language Processing tasks. However, understanding of their internal functioning is still insufficient and unsatisfactory. In order to better understand BERT and other Transformer-based models, we present a layer-wise analysis of BERT's hidden states. Unlike previous research, which mainly focuses on explaining Transformer models by their \hbox{attention} weights, we argue that hidden states contain equally valuable information. Specifically, our analysis focuses on models fine-tuned on the task of Question Answering (QA) as an example of a complex downstream task. We inspect how QA models transform token vectors in order to find the correct answer. To this end, we apply a set of general and QA-specific probing tasks that reveal the information stored in each representation layer. Our qualitative analysis of hidden state visualizations provides additional insights into BERT's reasoning process. Our results show that the transformations within BERT go through phases that are related to traditional pipeline tasks. The system can therefore implicitly incorporate task-specific information into its token representations. Furthermore, our analysis reveals that fine-tuning has little impact on the models' semantic abilities and that prediction errors can be recognized in the vector representations of even early layers.
\end{abstract}

\copyrightyear{2019}
\acmYear{2019}
\acmConference[CIKM '19]{The 28th ACM International Conference on Information and Knowledge Management}{November 3--7, 2019}{Beijing, China}
\acmBooktitle{The 28th ACM International Conference on Information and Knowledge Management (CIKM '19), November 3--7, 2019, Beijing, China}
\acmPrice{15.00}
\acmDOI{10.1145/3357384.3358028}
\acmISBN{978-1-4503-6976-3/19/11}

\keywords{neural networks, transformers, explainability, word representation, natural language processing, question answering}

\fancyhead{}

\maketitle

\section{Introduction}
In recent months, Transformer models have become more and more prevalent in the field of Natural Language Processing. Originally they became popular for their improvements over RNNs in Machine Translation \cite{attentionisall}. Now however, with the advent of large models and an
equally large amount of pre-training being done, they have proven
adept at solving many of the standard Natural Language Processing tasks. Main subject of this paper is BERT \cite{bert}, arguably the most popular of the recent Transformer models and the first to display significant improvements over previous state-of-the-art models in a number of different benchmarks and tasks.\\

\noindent\textbf{Problem of black box models}. Deep Learning models achieve increasingly impressive results across a number of different domains, whereas their application to real-world tasks has been moving somewhat more slowly. One major impediment lies in the lack of transparency, reliability and prediction guarantees in these largely black box models. 

While Transformers are commonly believed to be moderately interpretable through the inspection of their attention values, current research suggests that this may not always be the case \cite{jain2019attention}. This paper takes a different approach to the interpretation of said Transformer Networks. Instead of evaluating attention values, our approach examines the hidden states between encoder layers directly. There are multiple questions this paper will address:
\begin{enumerate}
    \item Do Transformers answer questions decompositionally, in a similar manner to humans?
    \item Do specific layers in a multi-layer Transformer network solve different tasks?
    \item How does fine-tuning influence the network's inner state?
    \item Can an evaluation of network layers help determine why and how a network failed to predict a correct answer?
\end{enumerate} 

We discuss these questions on the basis of fine-tuned models on standard QA datasets. We choose the task of Question Answering as an example of a complex downstream task that, as this paper will show, requires solving a multitude of other Natural Language Processing tasks. Additionally, it has been shown that other NLP tasks can be successfully framed as QA tasks \cite{decanlp}, therefore our analysis should translate to these tasks as well. While this work focuses on the BERT architecture, we perform preliminary tests on the small GPT-2 model \cite{gpt2} as well, which yield similar results.\\

\noindent\textbf{Contributions}. With the goal to improve understanding of \hbox{internal} workings of Transformers we present the following contributions:

First, we propose a layer-wise visualisation of token representations that reveals information about the internal state of Transformer networks. This visualisation can be used to expose wrong predictions even in earlier layers or to show which parts of the context the model considered as Supporting Facts.

Second, we apply a set of general NLP Probing Tasks and extend them by the QA-specific tasks of Question Type Classification and Supporting Fact Extraction. This way we can analyse the abilities within BERT’s layers and how they are impacted by fine-tuning.

Third, we show that BERT's transformations go through similar phases, even if fine-tuned on different tasks. Information about general language properties is encoded in earlier layers and implicitly used to solve the downstream task at hand in later layers.
\section{Related work}
\label{sec:relatedwork}

\noindent\textbf{Transformer Models}. Our analyses focus on BERT, which belongs to the group of Transformer networks, named after how representations are transformed throughout the network layers. We also partly include the more recent Transformer model GPT-2 \cite{gpt2}. This model represents OpenAI's improved version of GPT \cite{gpt} and while GPT-2 has not yet climbed leaderboards like BERT has, its larger versions have proven adept enough at the language modeling task, that Open-AI has decided not to release their pre-trained models. There are also other Transformer models of note, where a similar analysis might prove interesting in future work. Chief among them are the Universal Transformer \cite{universaltransformer} and TransformerXL \cite{transformerXL}, both of which aim to improve some of the flaws of the  Transformer architecture by adding a recurrent inductive bias.\\

\noindent\textbf{Interpretability and Probing}. Explainability and Interpretability of neural models have become an increasingly large field of research. While there are a multitude of ways to approach these topics \cite{exsurvey1, exsurvey2, exsurvey3}, we especially highlight relevant work in the area of research that builds and applies probing tasks and methodologies, post-hoc, to trained models.
There have been a number of recent advances on this topic. While the majority of the current works aim to create or apply more general purpose probing tasks \cite{senteval, probing1, probing3}, BERT specifically has also been probed in previous papers. 
\citet{BertProbing} proposes a novel "edge-probing" framework consisting of nine different probing tasks and applies it to the contextualized word embeddings of ELMo, BERT and GPT-1. Both semantic and syntactic information is probed, but only pre-trained models are studied, and not specifically fine-tuned ones. A similar analysis \cite{BertSyntactic} adds more probing tasks and addresses only the BERT architecture.

\citet{BertRanking} focus specifically on analysing BERT as a Ranking model. The authors probe attention values in different layers and measure performance for representations build from different BERT layers. Like \cite{BertProbing}, they only discuss pre-trained models.

There has also been work which studies models not through probing tasks but through qualitative visual analysis. \citet{visual1} offer a survey of different approaches, though limited to CNNs. \citet{visual2} explore phoneme recognition in DNNs by studying single node activations in the task of speech recognition. \citet{visual3} go one step further, by not only doing a qualitative analysis, but also training diagnostic classifiers to support their hypotheses. Finally, \citet{visual4} take a look at word vectors and the importance of some of their specific dimensions on both sequence tagging and classification tasks.

The most closely related previous work is proposed by  \citet{BertElmoLayerwise}. Here, the authors also perform a layer-wise analysis of BERT's token representations. However, their work solely focuses on probing pre-trained models and disregards models fine-tuned on downstream tasks. Furthermore, it limits the analysis to the general transferability of the network and does not analyze the specific phases that BERT goes through.

Additionally, our work is motivated by \citet{jain2019attention}. In their paper, the authors argue that attention, at least in some cases, is not well suited to solve the issues of explainability and interpretability. They do so both by constructing adversarial examples and by a comparison with more traditional explainability methods. In supporting this claim, we propose revisiting evaluating hidden states and token representations instead.  
\section{BERT under the Microscope}
We focus our analysis on fine-tuned BERT models. In order to understand which transformations the models apply to input tokens, we take two approaches: First, we analyse the transforming token vectors \textit{qualitatively} by examining their positions in vector space. Second, we probe their language abilities on QA-related tasks to examine our results \textit{quantitatively}.

\subsection{Analysis of Transformed Tokens}
The architecture of BERT and Transformer networks in general allows us to follow the transformations of each token throughout the network. We use this characteristic for an analysis of the changes that are being made to the tokens' representations in every layer.

We use the following approach for a qualitative analysis of these transformations:
We randomly select both correctly and falsely predicted samples from the test set of the respective dataset. For these samples we collect the hidden states from each layer while removing any padding. This results in the representation of each token throughout the model's layers.

The model can transform the vector space freely throughout its layers and we do not have references for semantic meanings of positions within these vector spaces. Therefore we consider distances between token vectors as indication for semantic relations.\\

\noindent\textbf{Dimensionality Reduction}. BERT's pre-trained models use vector dimensions of 1024 (large model) and 512 (base model). In order to visualize relations between tokens, we apply dimensionality reduction and fit the vectors into two-dimensional space. To that end we apply T-distributed Stochastic Neighbor Embedding (t-SNE) \cite{tsne}, Principal Component Analysis (PCA) \cite{pca} and Independent Component Analysis (ICA) \cite{ica} to vectors in each layer. As the results of PCA reveal the most distinct clusters for our data, we use it to present our findings.\\

\noindent\textbf{K-means Clustering}. In order to verify that clusters in 2D space represent the actual distribution in high-dimensional vector space, we additionally apply a k-means clustering \cite{kmeans}. We choose the number of clusters k in regard to the number of observed clusters in PCA, which vary over layers. The resulting clusters correspond with our observations in 2D space.

\begin{figure}[t!]
\includegraphics[width=0.42\textwidth]{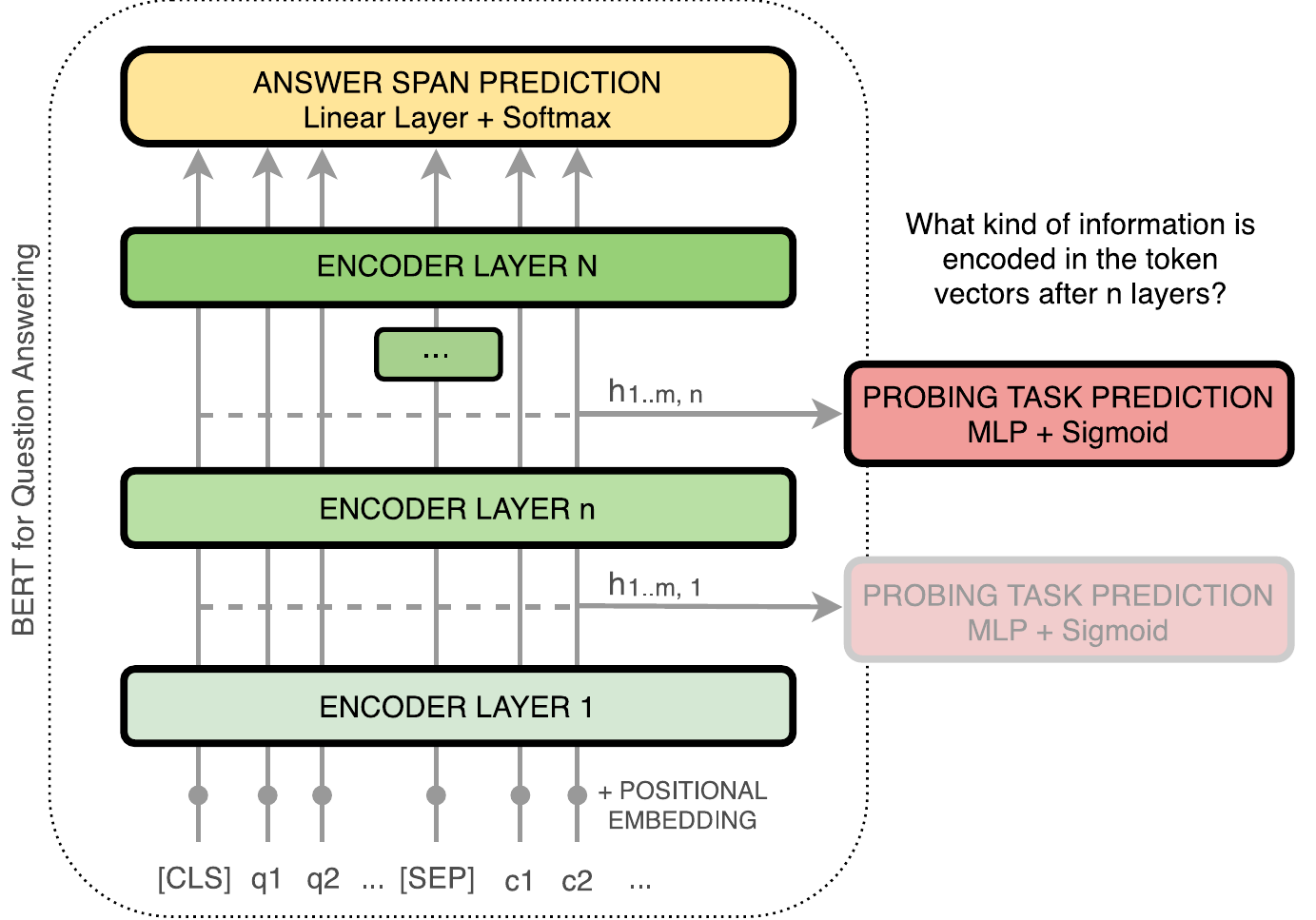}
\caption{Schematic overview of the BERT architecture and our probing setup. Question and context tokens are processed by N encoder blocks with a Positional Embedding added beforehand. The output of the last layer is fed into a span prediction head consisting of a Linear Layer and a Softmax. We use the hidden states of each layer as input to a set of probing tasks to examine the encoded information.}
\label{fig:bert_overview}
\end{figure}

\subsection{Probing BERT's Layers}
Our goal is to further understand the abilities of the model after each transformation. We therefore apply a set of semantic probing tasks to analyze which information is stored within the transformed tokens after each layer. We want to know whether specific layers are \textit{reserved} for specific tasks and how language information is maintained or forgotten by the model.

We use the principle of Edge Probing introduced by \citet{BertProbing}. Edge Probing translates core NLP tasks into classification tasks by focusing solely on their labeling part. This enables a standardized probing mechanism over a wide range of tasks.
We adopt the tasks Named Entity Labeling, Coreference Resolution and Relation Classification from the original paper as they are prerequisites for language understanding and reasoning \citep{babi}. We add tasks of Question Type Classification and Supporting Fact Identification due to their importance for Question Answering in particular.\footnote{The source code is available at: https://github.com/bvanaken/explain-BERT-QA}\\

\noindent\textbf{Named Entity Labeling}. Given a span of tokens the model has to predict the correct entity category. This is based on Named Entity Recognition but formulated as a Classification problem. The task was modeled by \cite{BertProbing}, annotations are based on the OntoNotes 5.0 corpus \citep{ontonotes} and contain 18 entity categories.\\

\noindent\textbf{Coreference Resolution}. The Coreference task requires the model  to predict whether two mentions within a text refer to the same entity. The task was built from the OntoNotes corpus and enhanced with negative samples by \cite{BertProbing}.\\

\noindent\textbf{Relation Classification}. In Relation Classification the model has to predict which relation type connects two known entities. The task was constructed by \cite{BertProbing} with samples taken from the SemEval 2010 Task 8 dataset consisting of English web text and nine directional relation types.\\

\noindent\textbf{Question Type Classification}. A fundamental part of answering a question is to correctly identify its question type. For this Edge Probing task we use the Question Classification dataset constructed by \citet{questionTypeDataset} based on the TREC-10 QA dataset \citep{trec10}. It includes 500 fine-grained types of questions within the larger groups of abbreviation, entity, description, human, location and numeric value. We use the whole question as input to the model with its question type as label.\\

\begin{table*}[t]
\begin{tabular}{|p{1.5cm} | p{7.5cm} | p{7.5cm}|}
\hline
& SQuAD & bAbI \\ \hline
Question & What is a common punishment in the UK and Ireland? & What is Emily afraid of? \\ \hline
Answer & \textbf{detention} & \textbf{cats} \\ \hline
Context & \pbox{7.5cm}{\vspace{1.5pt}\textbf{Currently detention is one of the most common punishments in schools in the United States, the UK, Ireland, Singapore and other countries.} It requires the pupil to remain in school at a given time in the school day (such as lunch, recess or after school); or even to attend school on a non-school day, e.g. "Saturday detention" held at some schools. During detention, students normally have to sit in a classroom and do work, write lines or a punishment essay, or sit quietly.\vspace{1.5pt}}
 & \pbox{7.5cm}{\textbf{Wolves are afraid of cats.}\\Sheep are afraid of wolves.\\Mice are afraid of sheep.\\Gertrude is a mouse.\\Jessica is a mouse.\\\textbf{Emily is a wolf.}\\Cats are afraid of sheep.\\Winona is a wolf.} \\ \hline
\end{tabular}
\caption{Samples from SQuAD dataset (left) and from Basic Deduction task (\#15) of the bAbI dataset (right). Supporting Facts are printed in bold. The SQuAD sample can be solved by word matching and entity resolution, while the bAbI sample requires a logical reasoning step and cannot be solved by simple word matching. Figures in the further analysis will use these examples where applicable.}
\label{table:squad-babi-examples}
\end{table*}

\noindent\textbf{Supporting Facts}. The extraction of Supporting Facts is essential for Question Answering tasks, especially in the multi-hop case. We examine what BERT's token transformations can tell us about the mechanism behind identifying important context parts.

To understand at which stage this distinction is done, we construct a probing task for identifying Supporting Facts. The model has to predict whether a sentence contains supporting facts regarding a specific question or whether it is irrelevant. Through this task we test the hypothesis that token representations contain information about their significance to the question. 

Both HotpotQA and bAbI contain information about sentence-wise Supporting Facts for each question. SQuAD does not require multi-hop reasoning, we therefore consider the sentence containing the answer phrase the Supporting Fact. We also exclude all QA-pairs that only contain one context sentence. We construct a different probing task for each dataset in order to check their task-specific ability to recognize relevant parts. All samples are labeled sentence-wise with true if they are a supporting fact or false otherwise.\\

\noindent\textbf{Probing Setup}. Analogue to the authors of \cite{BertProbing}, we embed input tokens for each probing task sample with our fine-tuned BERT model. Contrary to previous work, we do this for all layers (\(N=12\) for BERT-base and \(N=24\) for BERT-large), using only the output embedding from \(n\)-th layer at step \(n\). The concept of Edge Probing defines that only tokens of "labeled edges" (e.g. tokens of two related entities for Relation Classification) within a sample are considered for classification. These tokens are first pooled for a fixed-length representation and afterwards fed into a two-layer Multi-layer Perceptron (MLP) classifier, that predicts label-wise probability scores (e.g. for each type of relation).
A schematic overview of this setting is shown in Figure \ref{fig:bert_overview}.
We perform the same steps on pre-trained BERT-base and BERT-large models without any fine-tuning. This enables us to identify which abilities the model learns during pre-training or fine-tuning.

\section{Datasets and Models}
\label{sec:datasetsmodels}

\subsection{Datasets}
Our aim is to understand how BERT works on complex downstream tasks. Question Answering (QA) is one of such tasks that require a combination of multiple simpler tasks such as Coreference Resolution and Relation Modeling to arrive at the correct answer.
We take three current Question Answering datasets into account, namely SQUAD \citep{squad}, bAbI \citep{babi} and HotpotQA \citep{hotpot}. We intentionally choose three very different datasets to diversify the results of our analysis.\\

\noindent\textbf{SQuAD}. As one of the most popular QA tasks the SQuAD dataset contains ~100,000 natural question-answer pairs on ~500 Wikipedia articles. A new version of the dataset called SQuAD 2.0 \citep{squad2} additionally includes unanswerable questions. We use the previous version SQuAD 1.1 for our experiments to concentrate on the base task of span prediction. In 2018 an ensemble of fine-tuned BERT models has outperformed the Human Baseline on this task.
The dataset is characterised by questions that mainly require to resolve lexical and syntactic variations.\\

\noindent\textbf{HotpotQA}. This Multihop QA task contains ~112,000 natural question-answer pairs. The questions are especially designed to combine information from multiple parts of a context. 
We focus on the \textit{distractor}-task of HotpotQA, in which the context is composed of both supporting and distracting facts with an average size of 900 words. As the pre-trained BERT model is restricted to an input size of 512 tokens, we reduce the amount of distracting facts by a factor of 2.7. 
We also leave out yes/no-questions (7\% of questions) as they require additional specific architecture, diluting our analysis.\\

\noindent\textbf{bAbI}. The QA bAbI tasks are a set of artificial toy tasks developed to further understand the abilities of neural models. The 20 tasks require reasoning over multiple sentences (Multihop QA) and are modeled to include Positional Reasoning, Argument Relation Extraction and Coreference Resolution. The tasks strongly differ from the other QA tasks in their simplicity (e.g. vocabulary  size of ~230 and short contexts) and the artificial nature of sentences.

\begin{table}[]
\begin{tabular}{@{}lllll@{}}
\toprule
 & SQuAD &  HotpotQA Distr. & HotpotQA SP &   bAbI\\ \midrule
Baseline & 77.2 & 66.0  & 66.0 & 42.0 \\
BERT & 87.9 & 56.8 & 80.4 & 93.4 \\
GPT-2 & 74.9 & 54.0 & 64.6 & 99.9 \\ \bottomrule
\end{tabular}
\caption{Results from fine-tuning BERT on QA tasks. Baselines are: BIDAF \cite{bidaf} for SQuAD, the LSTM Baseline for bAbI from \cite{babi} and the HotpotQA baseline from \cite{hotpot} for the two Hotpot tasks.}
\label{table:model-results}
\end{table} 

\subsection{BERT and GPT-2}
In this section we briefly discuss the models our analysis is based on, BERT \cite{bert} and GPT-2 \cite{gpt2}.
Both of these models are Transformers that extend and improve on a number of different recent ideas. These include previous Transformer models \cite{attentionisall}\cite{gpt}, Semi-Supervised Sequence Learning \cite{semisuper}, ELMo \cite{elmo} and ULMFit \cite{ulmfit}. Both have a similar architecture, and they each represent one half of the original Encoder-Decoder Transformer \cite{attentionisall}. While GPT-2, like its predecessor, consists of only the decoder half, BERT uses a bidirectional variant of the original encoder. Each consists of a large number of Transformer blocks (12 for small GPT-2 and bert-base, 24 for bert-large), that in turn consist of a Self-Attention module, Feed Forward network, Layer Normalization and Dropout.
On top of these encoder stacks we add a Sequence Classification head for the bAbI dataset and a Span Prediction head for the other datasets. Figure \ref{fig:bert_overview} depicts how these models integrate into our probing setup.

\subsection{Applying BERT to Question Answering}
We base our training code on the Pytorch implementation of BERT available at \cite{pytorchBert}.
We use the publicly available pre-trained BERT models for our experiments. In particular, we study the monolingual models \textit{bert-base-uncased} and \textit{bert-large}. For GPT-2 the small model (117M Parameters) is used, as a larger model has not yet been released.
However, we do not apply these models directly, and instead fine-tune them on each of our datasets. \\

\noindent\textbf{Training Modalities}. Regarding hyperparameters, we tune the learning rate, batch size and learning rate scheduling according to a grid search 
and train each model for 5 epochs with evaluations on the development set every 1000 iterations. We then select the model of the best evaluation for further analysis. The input length chosen is 384 tokens for the bAbI and SQuAD tasks and the maximum of 512 tokens permitted by the pre-trained models' positional embedding for the HotpotQA tasks.
For bAbI we evaluate both models that are trained on a single bAbI task and also a multitask model, that was trained on the data of all 20 tasks. We further distinguish between two settings: Span prediction, which we include for better comparison with the other datasets, and Sequence Classification, which is the more common approach to bAbI. In order to make span prediction work, we append all possible answers to the end of the base context, since not all answers can be found in the context by default. 
For HotpotQA, we also distinguish between two tasks. In the \textit{HotpotQA Support Only} (SP) task, we use only the sentences labeled as Supporting Facts as the question context. This simplifies the task, but more importantly it reduces context length and increases our ability to distinguish token vectors. Our \textit{HotpotQA Distractor} task is closer to the original HotpotQA task. It includes distracting sentences in the context, but only enough to not exceed the 512 token limit.

%
%
%
%
%
%
\section{Results and Discussion}

\begin{figure}[]
\includegraphics[width=0.48\textwidth]{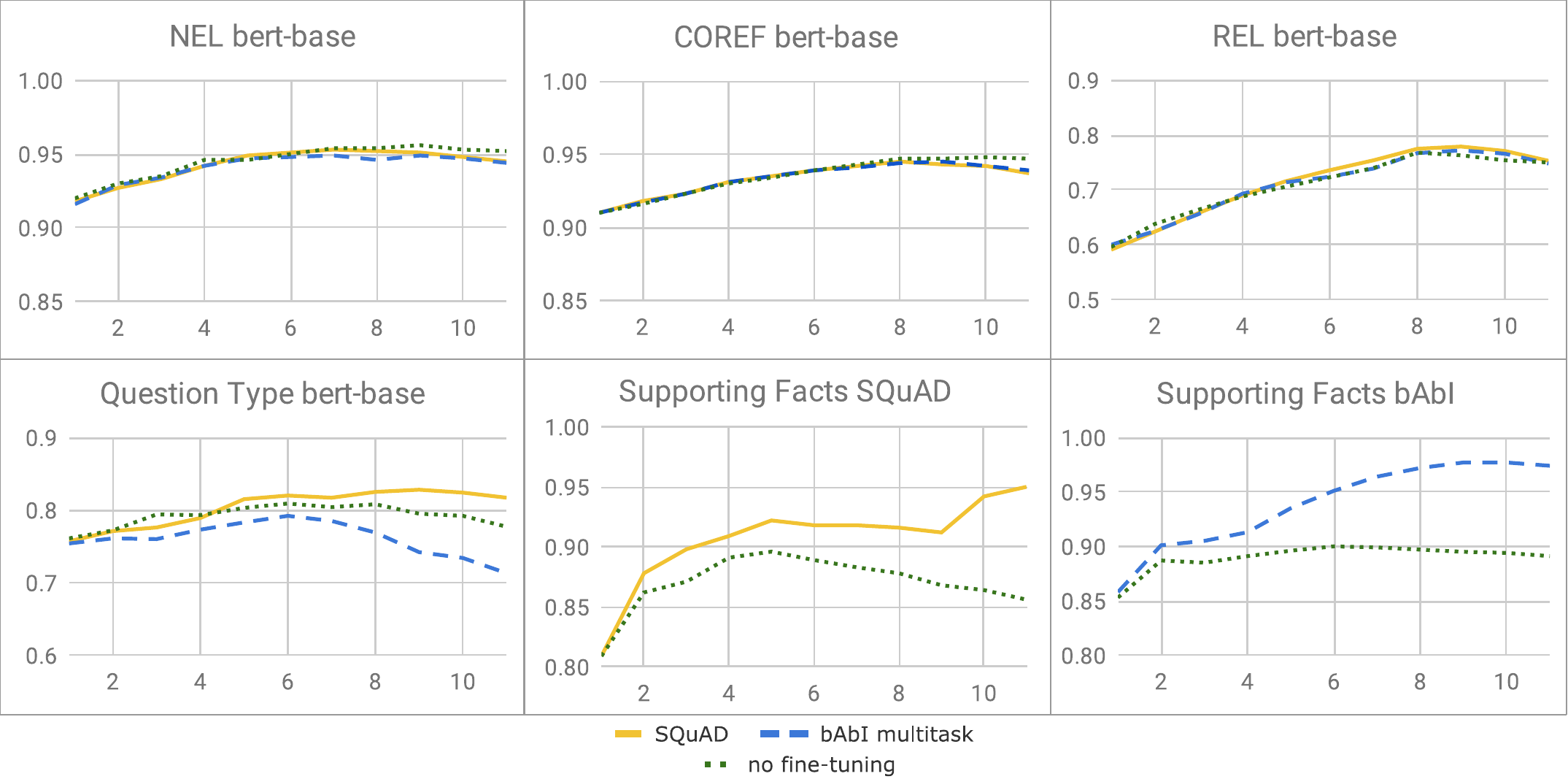}
\caption{Probing Task results of BERT-base models in macro averaged F1 (Y-axis) over all layers (X-axis). Fine-tuning barely affects accuracy on NEL, COREF and REL indicating that those tasks are already sufficiently covered by pre-training. Performances on the Question Type task shows its relevancy for solving SQuAD, whereas it is not required for the bAbI tasks and the information is lost.}
\label{fig:probing-bert-base}
\end{figure}

\begin{figure}[]
\includegraphics[width=0.48\textwidth]{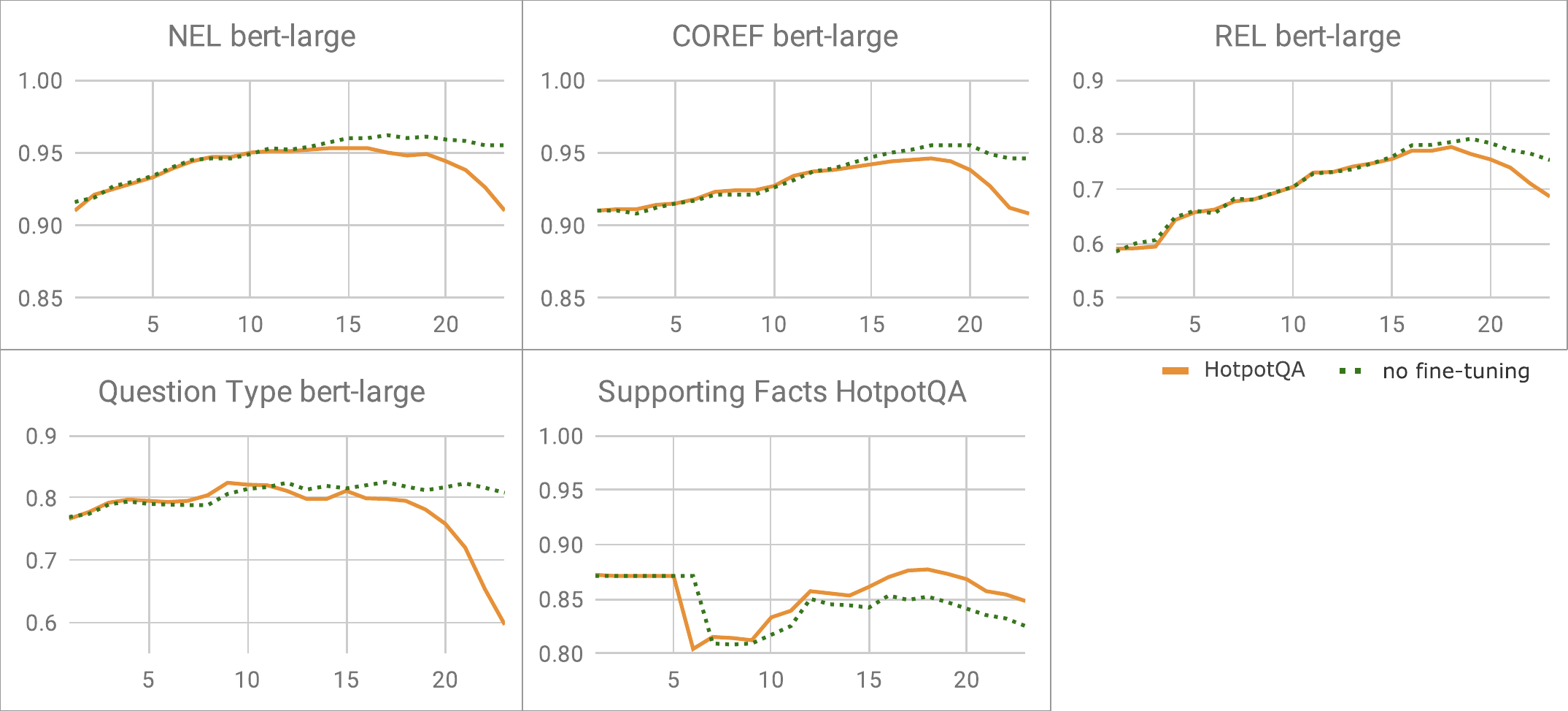}
\caption{Probing Task results of BERT-large models in macro averaged F1 (Y-axis) over all layers (X-axis). Performance of HotpotQA model is mostly equal to the model without fine-tuning, but information is dropped in last layers in order to fit the Answer Selection task.}
\label{fig:probing-bert-large}
\end{figure}

\noindent\textbf{Training Results}. Table \ref{table:model-results} shows the evaluation results of our best models. Accuracy on the SQuAD task is close to human performance, indicating that the model can fulfill all sub-tasks required to answer SQuAD's questions. As expected the tasks derived from HotpotQA prove much more challenging, with the distractor setting being the most difficult to solve. Unsurprisingly too, bAbI was easily solved by both BERT and GPT-2. While GPT-2 performs significantly worse in the more difficult tasks of SQuAD and HotpotQA, it does considerably better on bAbi reducing the validation error to nearly 0.
Most of BERT's error in the bAbI multi-task setting comes from tasks 17 and 19. Both of these tasks require positional or geometric reasoning, thus it is reasonable to assume that this is a skill where GPT-2 improves on BERT's reasoning capabilities.\\

\noindent\textbf{Presentation of Analysis Results}. The qualitative analysis of vector transformations reveals a range of recurring patterns. In the following, we present these patterns by two representative samples from the SQuAD and bAbI task dataset described in Table \ref{table:squad-babi-examples}. Examples from HotpotQA can be found in the supplementary material as they require more space due to the larger context.

Results from probing tasks are displayed in Figures \ref{fig:probing-bert-base} and \ref{fig:probing-bert-large}. We compare results in macro-averaged F1 over all network layers. Figure \ref{fig:probing-bert-base} shows results from three models of BERT-base with twelve layers: Fine-tuned on SQuAD,on bAbI tasks and without fine-tuning. Figure \ref{fig:probing-bert-large} reports results of two models based on BERT-large with 24 layers: Fine-tuned on HotpotQA and without fine-tuning.

\begin{figure*}[htp]
  \centering
\label{fig:phases-squad}
  \subfloat[SQuAD Phase 1: Semantic Clustering. We observe a topical cluster with\newline 'school'-related and another with 'country'-related tokens.]{\label{fig:squad-example-first-phase}\includegraphics[width=0.5\textwidth]{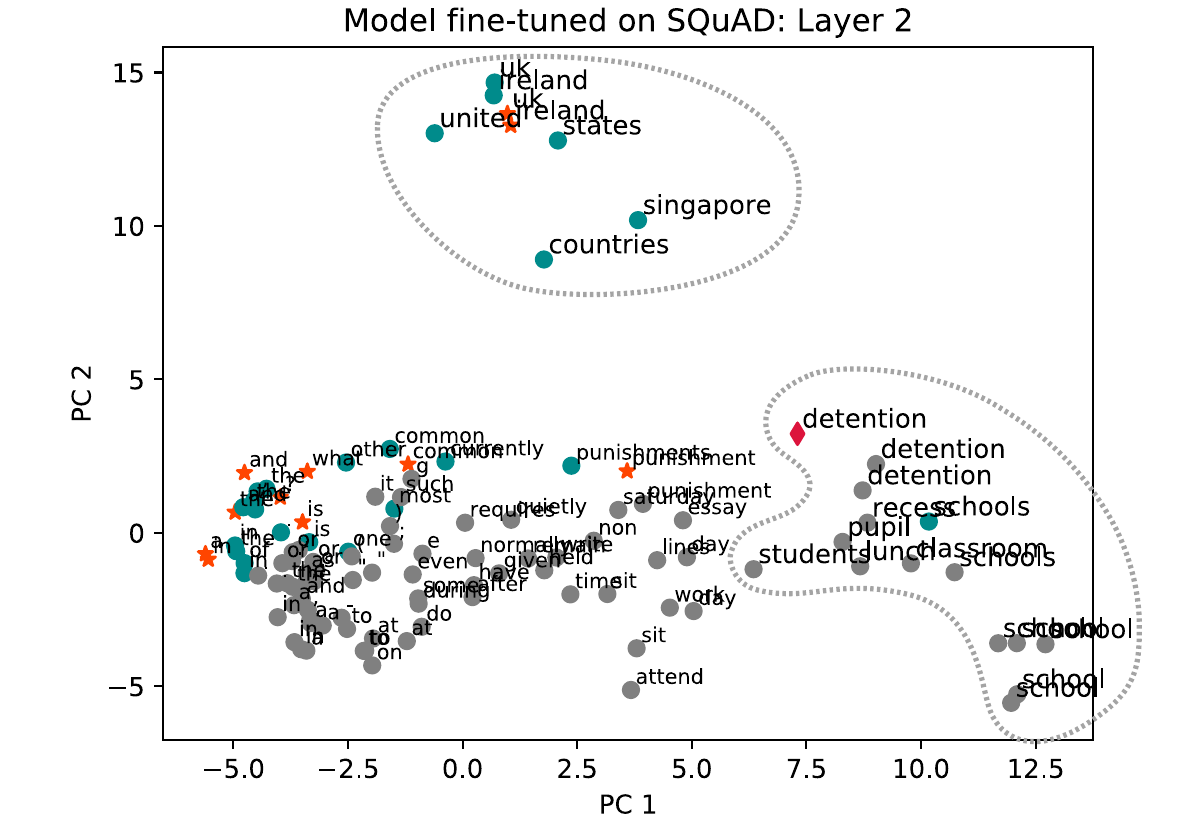}}
  \subfloat[SQuAD Phase 2: Entity Matching. The marked cluster contains matched tokens 'detention', 'schools' and the countries that are applying this practice. ]{\label{fig:squad-example-second-phase}\includegraphics[width=0.5\textwidth]{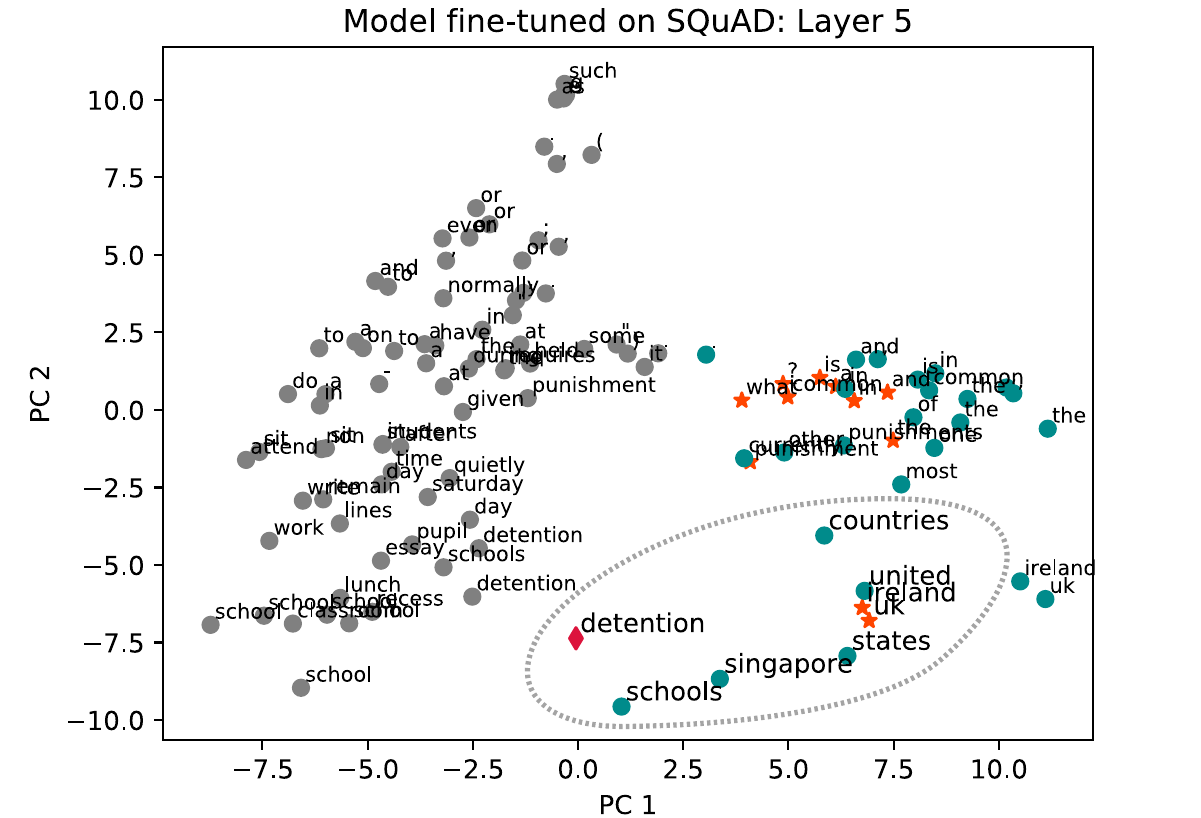}}
  \\
  \subfloat[SQuAD Phase 3: Question-Fact Matching. The question tokens form a\newline cluster with the Supporting Fact tokens. ]{\label{fig:squad-example-third-phase}\includegraphics[width=0.5\textwidth]{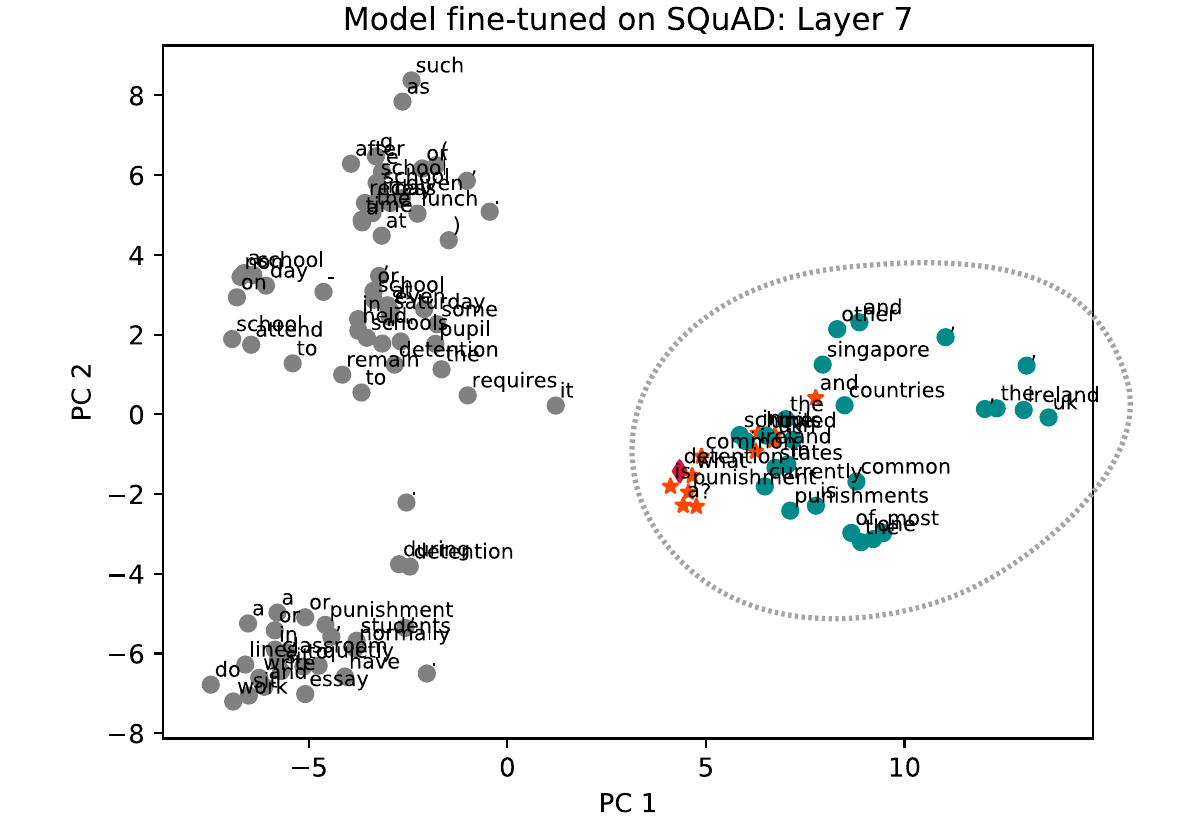}}
  \subfloat[SQuAD Phase 4: Answer Extraction. The answer token 'detention' is separated from other tokens.]{\label{fig:squad-example-fourth-phase}\includegraphics[width=0.5\textwidth]{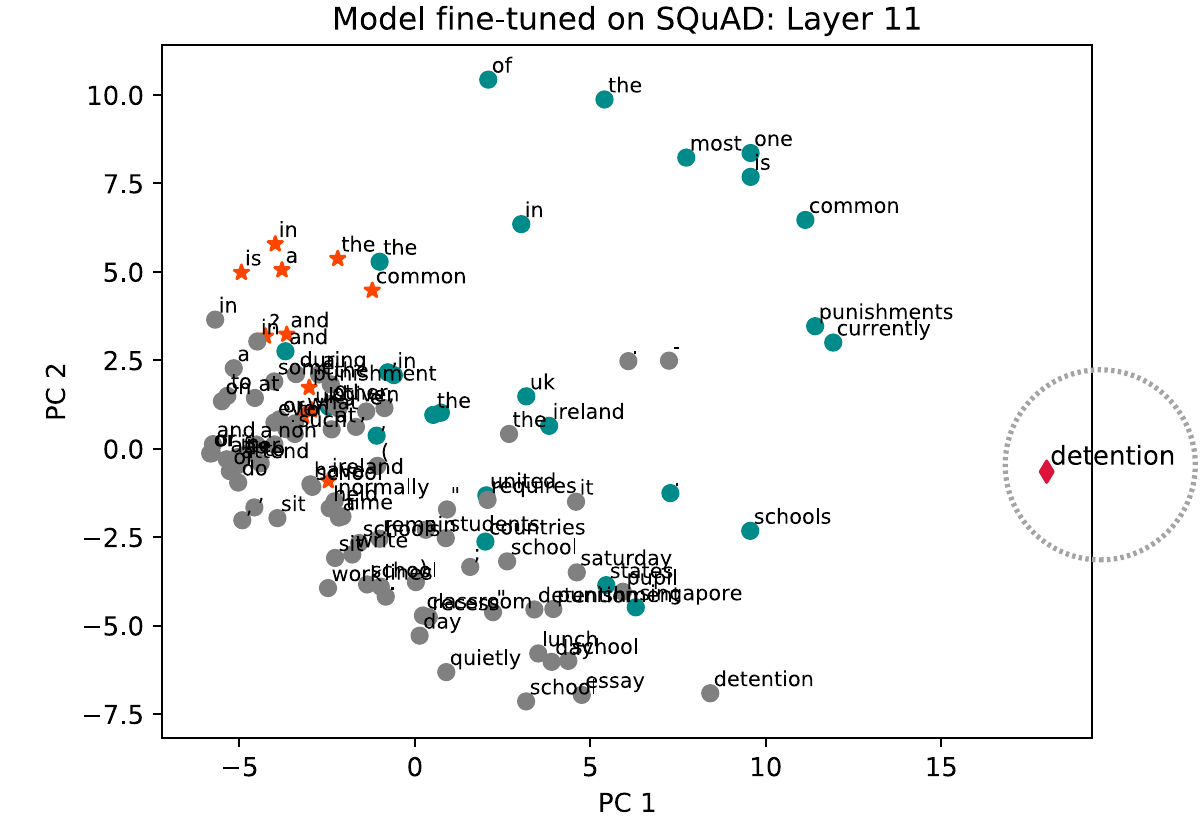}}
  \caption{BERT's Transformation Phases for the SQuAD example from Table \ref{table:squad-babi-examples}. Answer token: Red diamond-shaped. Question Tokens: Orange star-shaped. Supporting Fact tokens: Dark Cyan. Prominent clusters are circled. The model passes through different phases in order to find the answer token, which is extracted in the last layer (\#11).}
  \vspace{10pt}
\end{figure*}

\begin{figure*}[htp]
  \centering
\label{fig:phases-babi}
  \subfloat[bAbI Phase 1: Semantic Clustering. Names and animals are clustered.]{\label{fig:babi-example-first-phase}\includegraphics[width=0.5\textwidth]{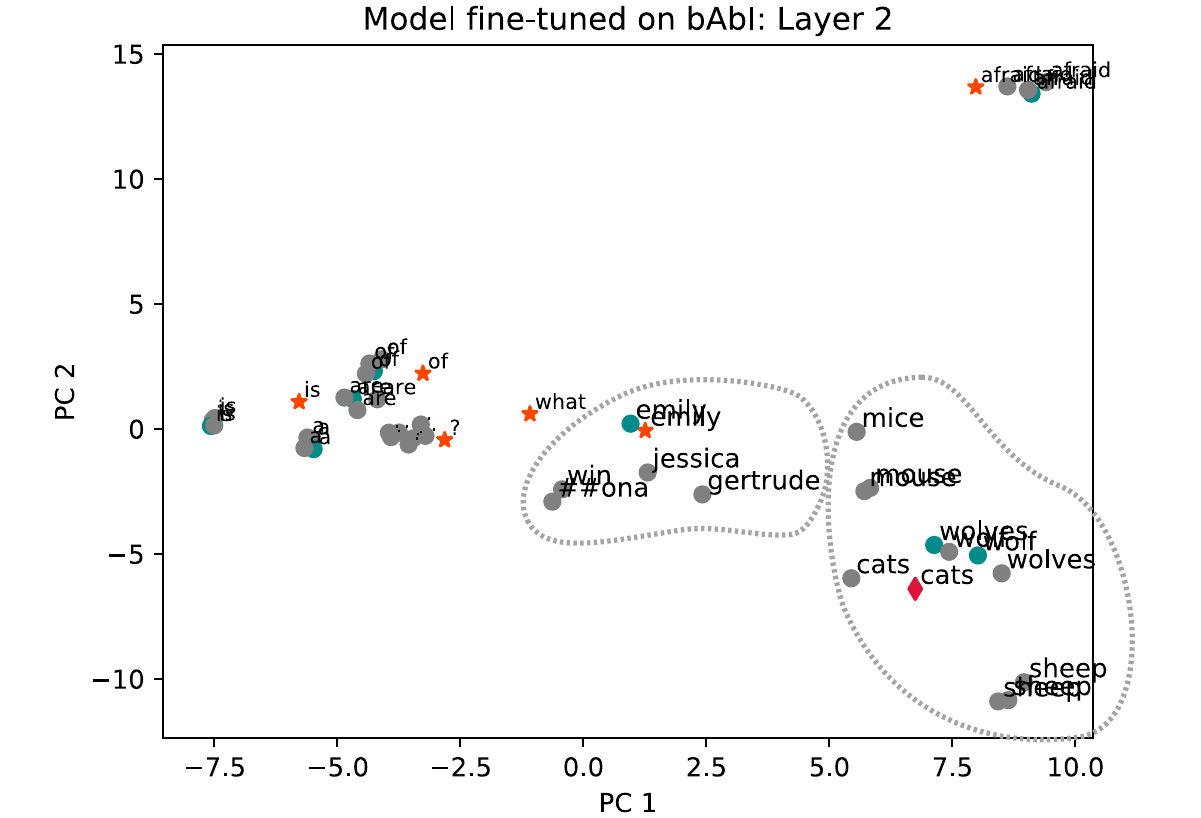}}
  \subfloat[bAbI Phase 2: Entity Matching. The determining relation between the entities 'Emily' and 'Wolf' is resolved in a cluster.]{\label{fig:babi-example-second-phase}\includegraphics[width=0.5\textwidth]{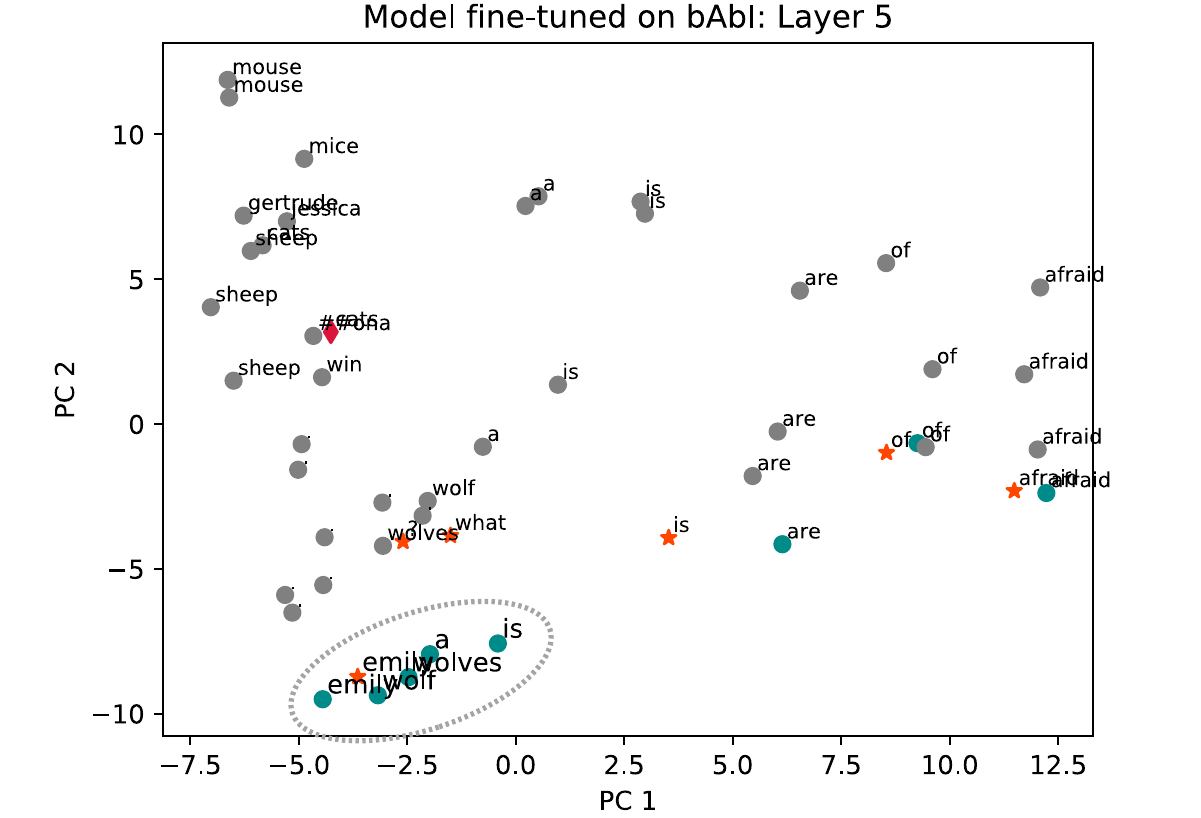}}
  \\
    \subfloat[bAbI Phase 3: Question-Fact Matching. In this case the question tokens\newline match with a subset of Supporting Facts ('Wolves are afraid of cats'). The\newline subset is decisive of the answer.]{\label{fig:babi-example-third-phase}\includegraphics[width=0.5\textwidth]{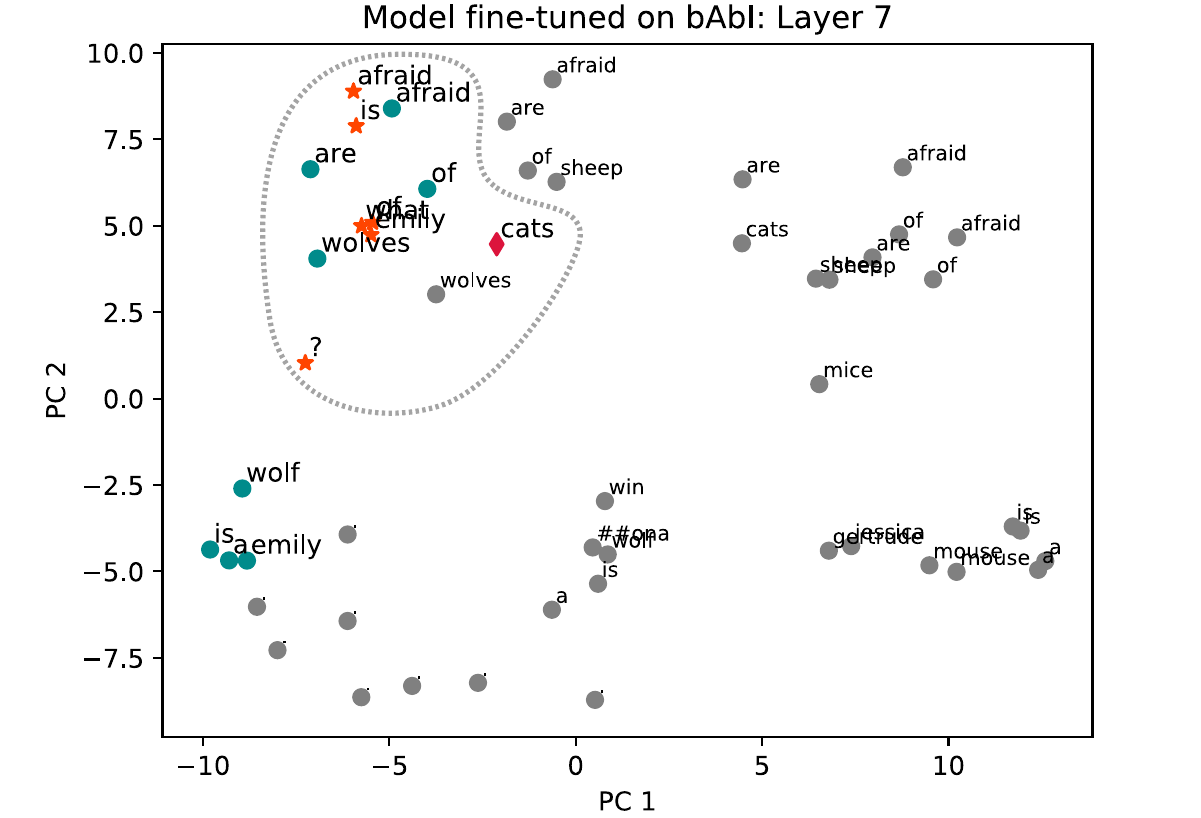}}
  \subfloat[bAbI Phase 4: Answer Extraction. The answer token 'cats' is separated from other tokens. ]{\label{fig:babi-example-fourth-phase}\includegraphics[width=0.5\textwidth]{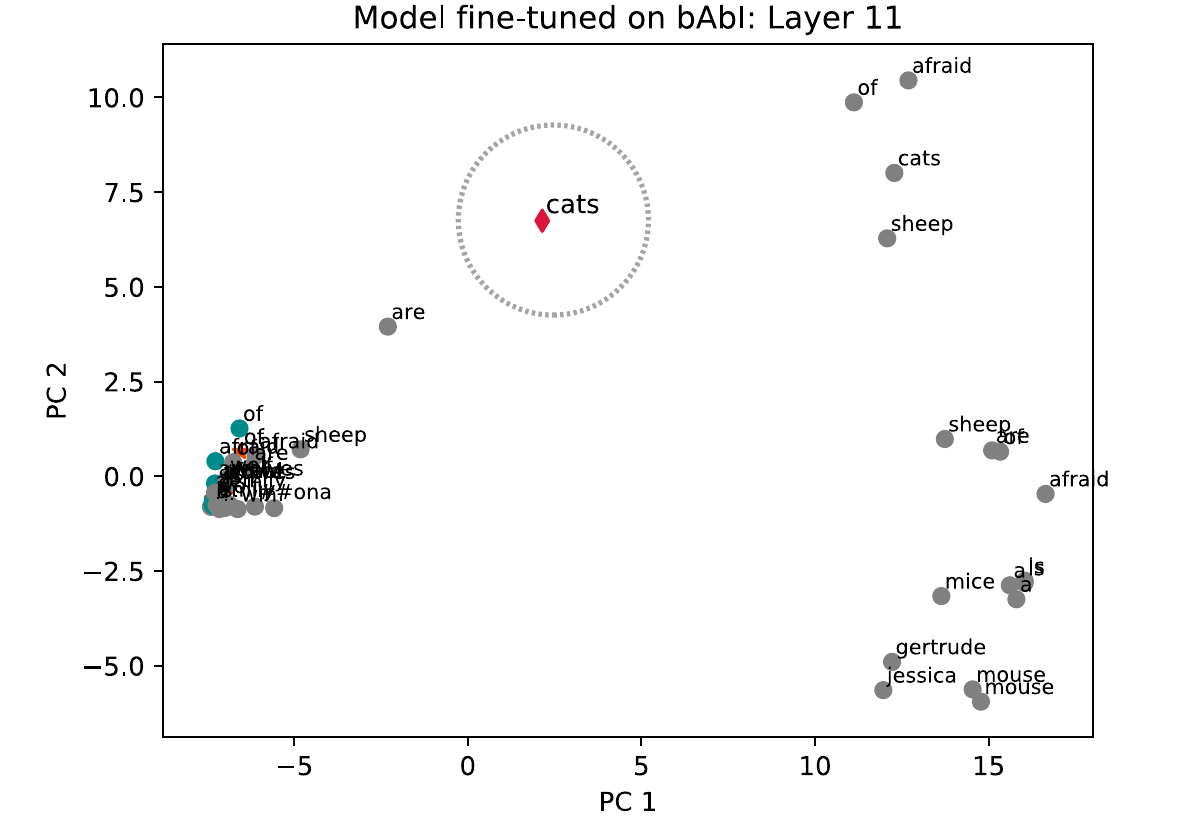}}
  \caption{BERT's Transformation Phases for the bAbI example from Table \ref{table:squad-babi-examples}. The phases are equal to what we observe in SQuAD and HotpotQA samples: The formed clusters in the first layers show general language abilities, while the last layers are more task-specific.}
\end{figure*}

\subsection{Phases of BERT's Transformations}
The PCA representations of tokens in different layers suggest that the model is going through multiple phases while answering a question. We observe these phases in all three selected QA tasks despite their diversity. These findings are supported by results of the applied probing tasks. We present the four phases in the following paragraphs and describe how our experimental results are linked. \\

\noindent\textbf{(1) Semantic Clustering}. Early layers within the BERT-based models group tokens into topical clusters.  Figures \ref{fig:squad-example-first-phase} and \ref{fig:babi-example-first-phase} reveal this behaviour and show the second layer of each model. Resulting vector spaces are similar in nature to embedding spaces from e.g. Word2Vec \cite{word2vec} and hold little task-specific information. Therefore, these initial layers reach low accuracy on semantic probing tasks, as shown in Figures \ref{fig:probing-bert-base} and \ref{fig:probing-bert-large}. BERT's early layers can be seen as an implicit replacement of embedding layers common in neural network architectures.\\

\noindent\textbf{(2) Connecting Entities with Mentions and Attributes}. In the middle layers of the observed networks we see clusters of entities that are less connected by their topical similarity. Rather, they are connected by their relation within a certain input context. These task-specific clusters appear to already include a filtering of question-relevant entities. Figure \ref{fig:squad-example-second-phase} shows a cluster with words like \textit{countries, schools, detention} and country names, in which 'detention' is a common practice in schools. This cluster helps to solve the question \textit{"What is a common punishment in the UK and Ireland?"}. Another question-related cluster is shown in Figure \ref{fig:babi-example-second-phase}. The main challenge within this sample is to identify the two facts that \textit{Emily is a wolf} and \textit{Wolves are afraid of cats}. The highlighted cluster implies that \textit{Emily} has been recognized as a relevant entity that holds a relation to the entity \textit{Wolf}. The cluster also contains similar entity mentions e.g. the plural form \textit{Wolves}. We observe analogous clusters in the HotpotQA model, which includes more cases of coreferences.

\begin{figure*}[t]
  \centering
  \includegraphics[width=0.95\textwidth]{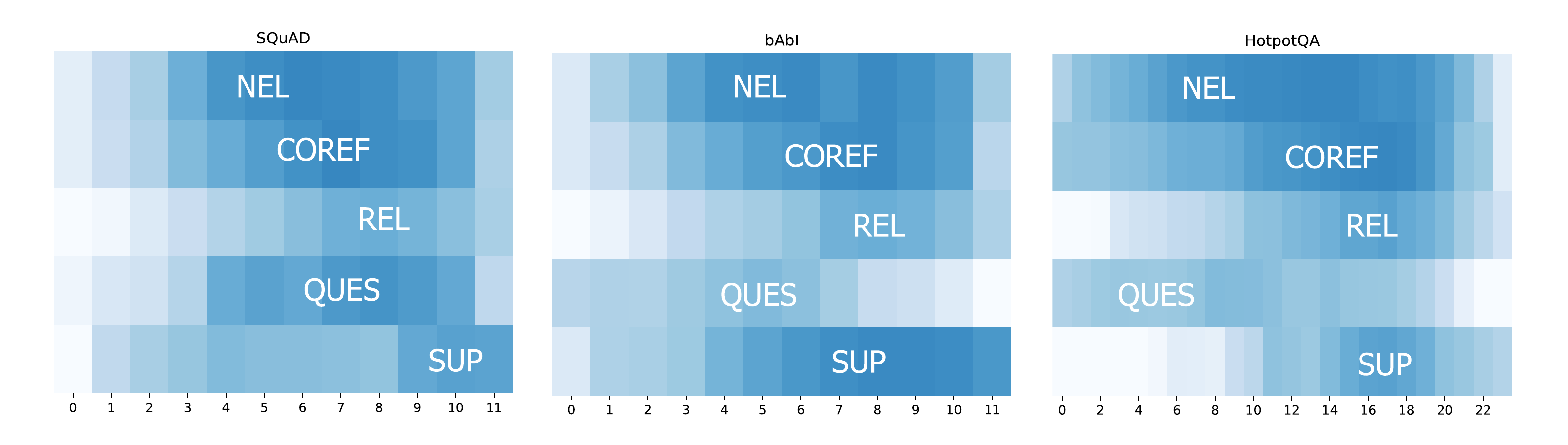}
  \caption{Phases of BERT's language abilities. Higher saturation denotes higher accuracy on probing tasks. Values are normalized over tasks on the Y-axis. X-axis depicts layers of BERT. NEL: Named Entity Labeling, COREF: Coreference Resolution, REL: Relation Classification, QUES: Question Type Classification, SUP: Supporting Fact Extraction. All three tasks exhibit similar patterns, except from QUES, which is solved earlier by the HotpotQA model based on BERT-large. NEL is solved first, while performance on COREF and REL peaks in later layers. Distinction of important facts (SUP) happens within the last layers.}
\label{fig:phases-bert-base}
\end{figure*}

The probing results support these observations. The model's ability to recognize entities (Named Entity Labeling), to identify their mentions (Coreference Resolution) and to find relations (Relation Recognition) improves until higher network layers. Figure \ref{fig:phases-bert-base} visualizes these abilities. Information about Named Entities is learned first, whereas recognizing coreferences or relations are more difficult tasks and require input from additional layers until the model's performance peaks. These patterns are equally observed in the results from BERT-base models and BERT-large models.\\

\noindent\textbf{(3) Matching Questions with Supporting Facts}. Identifying relevant parts of the context is crucial for QA and Information Retrieval in general. In traditional pipeline models this step is often achieved by filtering context parts based on their similarity to the question \cite{speechandlanguageprocessing}. 
We observe that BERT models perform a comparable step by transforming the tokens so that question tokens are matched onto relevant context tokens. Figures \ref{fig:squad-example-third-phase} and \ref{fig:babi-example-third-phase} show two examples in which the model transforms the token representation of question and Supporting Facts into the same area of the vector space. Some samples show this behaviour in lower layers. However, results from our probing tasks show that the models hold the strongest ability to distinguish relevant from irrelevant information wrt. the question in their higher layers. \hbox{Figure \ref{fig:probing-bert-base}} demonstrates how the performance for this task increases over successive layers for SQuAD and bAbI. Performance of the fine-tuned HotpotQA model in Figure \ref{fig:probing-bert-large} is less distinct from the model without fine-tuning and does not reach high accuracy.\footnote{Note that the model only predicts the majority class in the first five layers and thereby reaches a decent accuracy without really solving the task.} This inability indicates why the BERT model does not perform well on this dataset as it is not able to identify the correct Supporting Facts.

The vector representations enable us to tell which facts a model considered important (and therefore matched with the question). This  helps retracing decisions and makes the model more \hbox{transparent}.\\
    
\noindent\textbf{(4) Answer Extraction}. In the last network layers we see that the model dissolves most of the previous clusters. Here, the model separates the correct answer tokens, and sometimes other possible candidates, from the rest of the tokens. The remaining tokens form one or multiple homogeneous clusters. The vector representation at this point is largely task-specific and learned during fine-tuning. This becomes visible through the performance drop in general NLP probing tasks, visualized in Figure \ref{fig:phases-bert-base}. We especially observe this loss of information in last-layer representations in the large BERT-model fine-tuned on HotpotQA, as shown in Figure \ref{fig:probing-bert-large}. While the model without fine-tuning still performs well on tasks like NEL or COREF, the fine-tuned model loses this ability.\\

\noindent\textbf{Analogies to Human Reasoning}. The phases of answering questions can be compared to the human reasoning process,  including decomposition of input into parts \cite{fuzzy}. The first phase of semantic clustering represents our basic knowledge of language and the second phase how a human reader builds relations between parts of the context to connect information needed for answering a question. Separation of important from irrelevant information (phase 3) and grouping of potential answer candidates (phase 4) are also known from human reasoning. However, the order of these steps might differ from the human abstraction. One major difference is that while humans read sequentially, BERT can see all parts of the input at once. Thereby it is able to run multiple processes and phases concurrently depending on the task at hand. Figure \ref{fig:phases-bert-base} shows how the tasks overlap during the answering process.

\subsection{Comparison to GPT-2}
In this section we compare our insights from the BERT models to the GPT-2 model. We focus on the qualitative analysis of token representations and leave the application of probing tasks for future work. One major difference between GPT-2's and BERT's hidden states is that GPT-2 seems to give particular attention to the first token of a sequence. While in our QA setup this is often the question word, this also happens in cases where it is not. During dimensionality reduction this results in a separation of two clusters, namely the first token and all the rest. This problem holds true for all layers of GPT-2 except for the Embedding Layer, the first Transformer block and the last one. For this reason we mask the first token during dimensionality reduction in further analysis.

\begin{figure}[t]
  \includegraphics[width=0.5\textwidth]{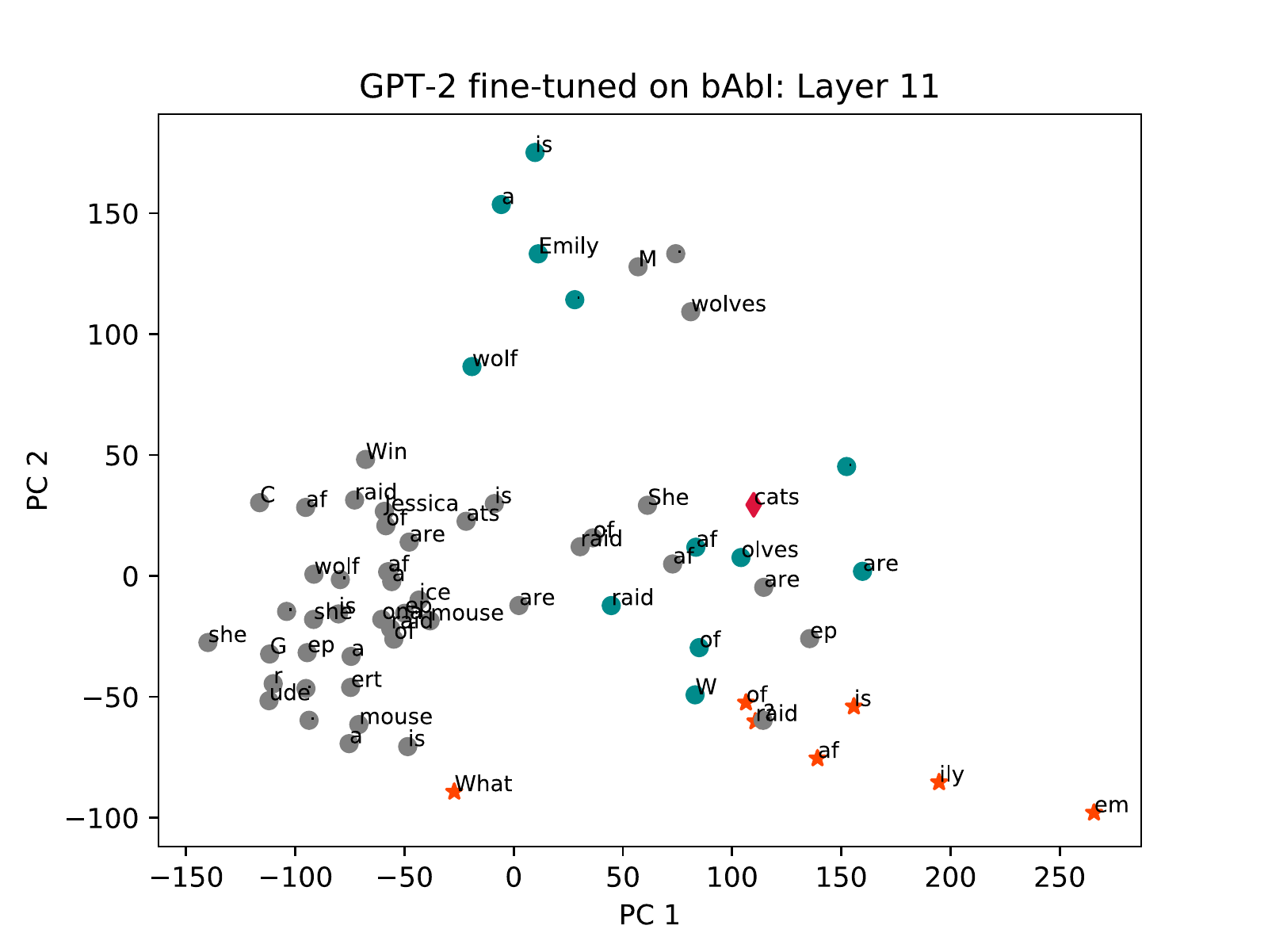}
  \caption{bAbI Example of the Answer Extraction phase in GPT-2. Both the question and Supporting Fact are extracted, but the correct answer is not fully separated as in BERT's last layers. Also a potential candidate Supporting Fact in "Sheep are afraid of Wolves" is separated as well.}
\label{fig:gpt2-example}
\end{figure}

Figure \ref{fig:gpt2-example} shows an example of the last layer's hidden state for our bAbI example. Like BERT, GPT-2 also separates the relevant Supporting Facts and the question in the vector space. Additionally, GPT-2 extracts another sentence, which is not a Supporting Fact, but is similar in meaning and semantics. In contrast to BERT, the correct answer "cats" is not particularly separated and instead simply left as part of its sentence. These findings in GPT-2 suggest that our analysis extends beyond the BERT architecture and hold true for other Transformer networks as well. Our future work will include more probing tasks to confirm this initial observation. 

\subsection{Additional Findings}
\begin{figure}[t!]
\includegraphics[width=0.5\textwidth]{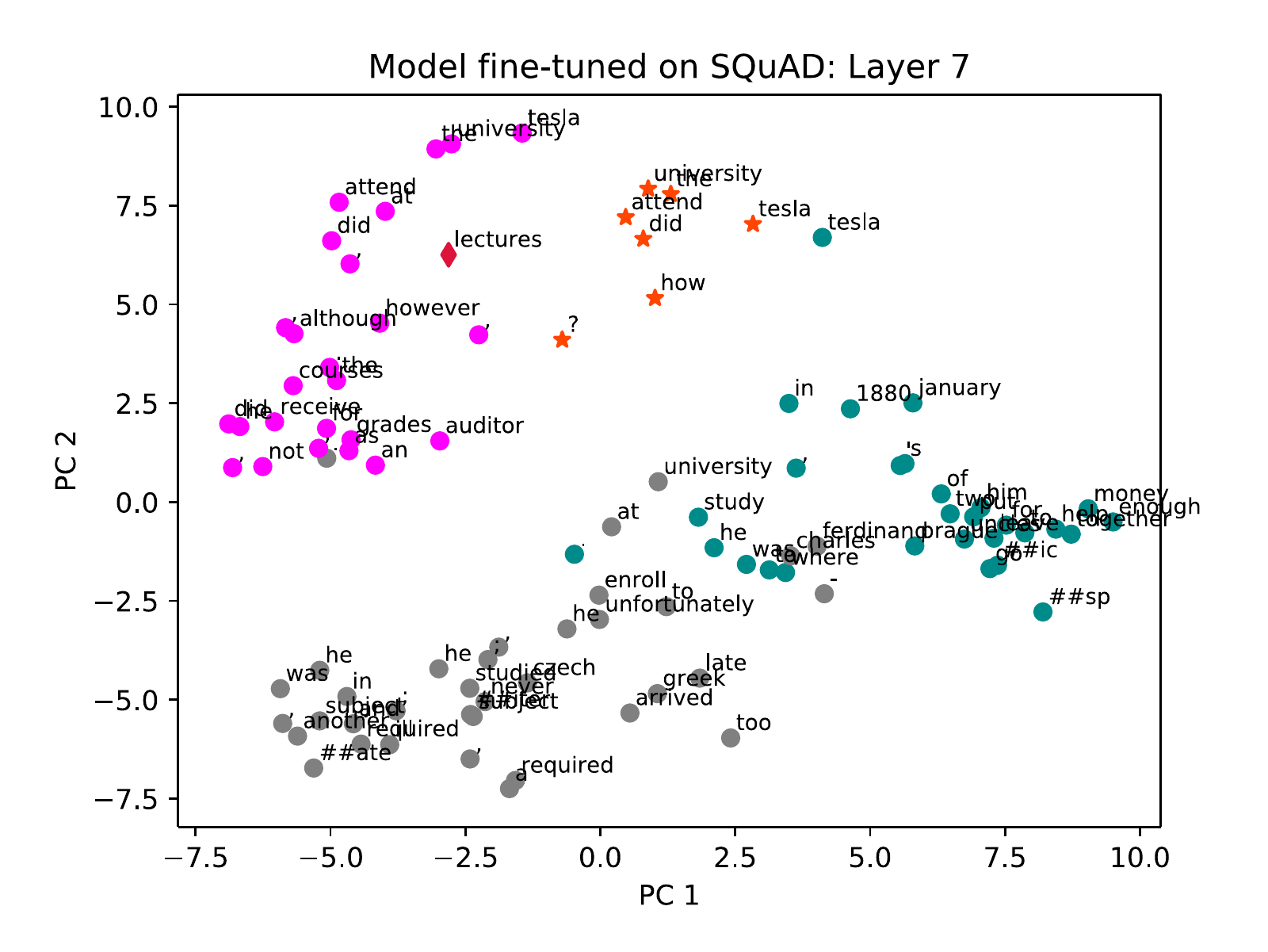}
\caption{BERT SQuAD example of a falsely selected answer based on the matching of the wrong Supporting Fact. The predicted answer 'lectures' is matched onto the question as a part of this incorrect fact (magenta), while the actual Supporting Fact (cyan) is not particularly separated.}
\label{fig:wrong_candidate}
\end{figure}

\textbf{Observation of Failure States}. One important aspect of explainable Neural Networks is to answer the questions of when, why, and how the network fails. Our visualizations are not only able to show such failure states, but even the rough difficulty of a specific task can be discerned by a glance at the hidden state representations. While for correct predictions the transformations run through the phases discussed in previous sections, for wrong predictions there exist two possibilities: If a candidate answer was found that the network has a reasonable amount of confidence in, the phases will look very similar to a correct prediction, but now centering on the wrong answer. Inspecting early layers in this case can give insights towards the reason why the wrong candidate was chosen, e.g. wrong Supporting Fact selected, misresolution of coreferences etc. An example of this is shown in Figure \ref{fig:wrong_candidate}, where a wrong answer is based on the fact that the wrong Supporting Fact was matched with the question in early layers.

If network confidence is low however, which is often the case when the predicted answer is far from the actual answer, the transformations do not go through the phases discussed earlier. The vector space is still transformed in each layer, but tokens are mostly kept in a single homogeneous cluster. In some cases, especially when the confidence of the network is low, the network maintains Phase (1), 'Semantic Clustering' analogue to Word2Vec, even in later layers. An example is depicted in the supplementary material.\\

\noindent\textbf{Impact of Fine-tuning}. Figures \ref{fig:probing-bert-base} and \ref{fig:probing-bert-large} show how little impact fine-tuning has on the core NLP abilities of the model. The pre-trained model already holds sufficient information about words and their relations, which is the reason it works well in multiple downstream tasks. Fine-tuning only applies small weight changes and forces the model to forget some information in order to fit specific tasks. However, the model does not forget much of the previously learned encoding when fitting the QA task, which indicates why the Transfer Learning approach proves successful.\\

\begin{figure}[t!]
\includegraphics[width=0.5\textwidth]{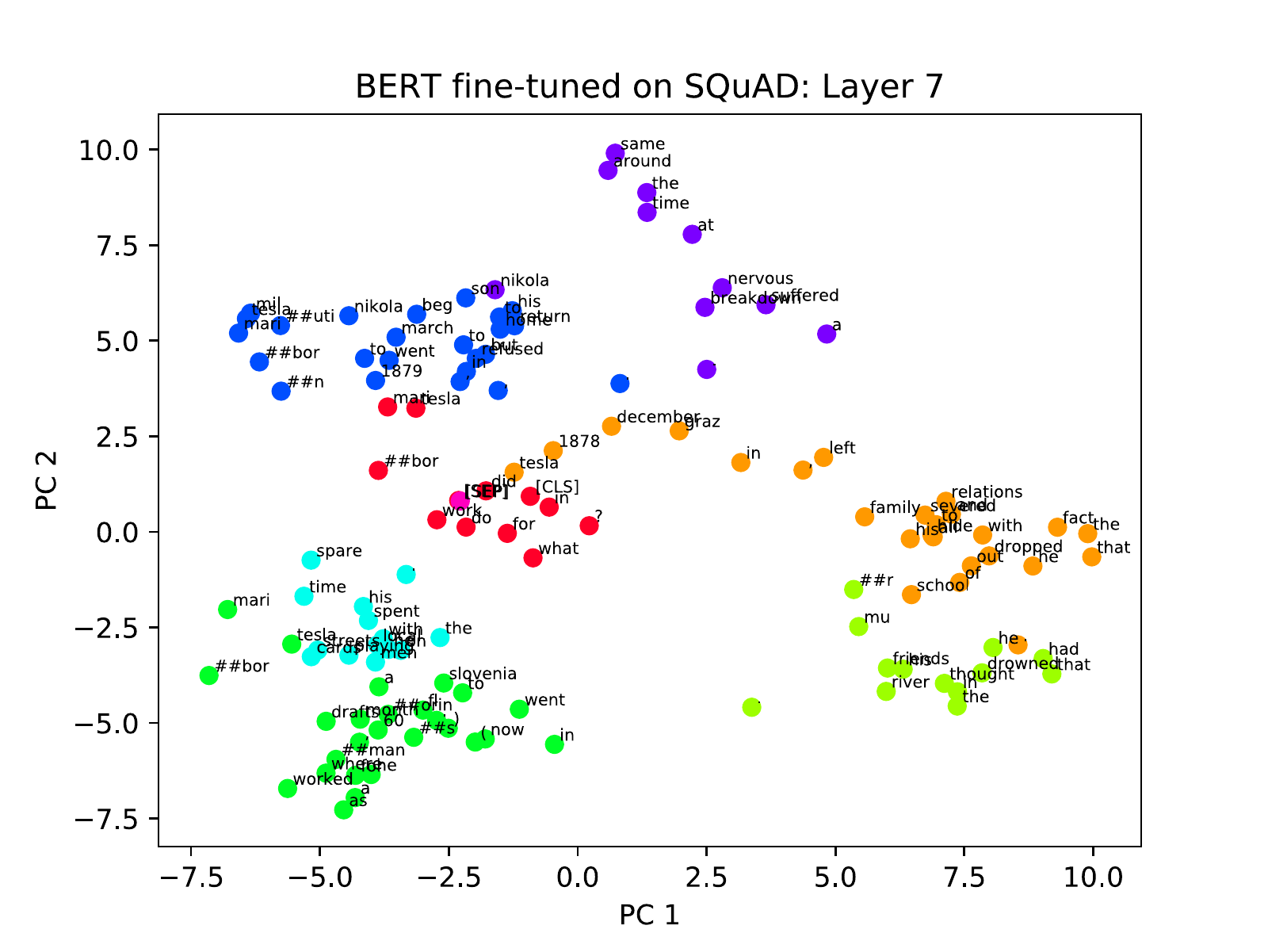}
\caption{BERT SQuAD example Layer 7. Tokens are color-coded by sentence. This visualization shows that tokens are clustered by their original sentence membership suggesting far reaching importance of the positional embedding.}
\label{fig:pos_embedding}
\end{figure}

\noindent\textbf{Maintained Positional Embedding}. It is well known that the positional embedding is a very important factor in the performance of Transformer networks. It solves one major problem that Transformers have in comparison with RNNs, that they lack sequential information \cite{attentionisall}. Our visualizations support this importance and show that even though the positional embedding is only added once before the first layer, its effects are maintained even into very late layers depending on the task. Figure \ref{fig:pos_embedding} demonstrates this behavior on the SQuAD dataset.\\

\noindent\textbf{Abilities to resolve Question Type}. The performance curves regarding the Question Type probing task illustrate another interesting result. Figure \ref{fig:probing-bert-base} demonstrates that the model fine-tuned on SQuAD outperforms the base model from layer 5 onwards. This indicates the relevancy of resolving the question type for the SQuAD task, which leads to an improved ability after fine-tuning. The opposite is the case for the model fine-tuned on the bAbI tasks, which loses part of its ability to distinguish question types during fine-tuning. This is likely caused by the static structure of bAbI samples, in which the answer candidates can be recognized by sentence structure and occurring word patterns rather than by the question type. Surprisingly, we see that the model fine-tuned on HotpotQA does not outperform the model without fine-tuning in Figure \ref{fig:probing-bert-large}. Both models can solve the task in earlier layers, which suggests that the ability to recognize question types is pre-trained in BERT-large.
\section{Conclusion and Future Work}
\label{sec:conclusion}
Our work reveals important findings about the inner functioning of Transformer networks. The impact of these findings and how future work can build upon them is described in the following:\\

\noindent\textbf{Interpretability}. The qualitative analysis of token vectors reveals that there is \hbox{indeed} interpretable information stored within the hidden states of Transformer models. This information can be used to identify misclassified examples and model weaknesses. It also provides clues about which parts of the context the model considered important for answering a question - a crucial part of decision legitimisation. We leave the development of methods to further process this information for future work.\\

\noindent\textbf{Transferability}. We further show that lower layers might be more applicable to certain problems than later ones. For a Transfer Learning task, this means layer depth should be chosen individually depending on the task at hand. We also suggest further work regarding skip connections in Transformer layers to examine whether direct information transfer between non-adjacent layers (that solve different tasks) can be of advantage.\\

\noindent\textbf{Modularity}. Our findings support the hypothesis that not only do different phases exist in Transformer networks, but that specific layers seem to solve different problems. This hints at a \hbox{modularity} that can potentially be exploited in the training process. For example, it could be beneficial to fit parts of the network to specific tasks in pre-training, instead of using an end-to-end language model task.\\

Our work aims towards revealing some of the internal processes within Transformer-based models. We suggest to direct further research at thoroughly understanding state-of-the-art models and the way they solve downstream tasks, in order to improve on them.

\begin{acks}
Our work is funded by the European Unions Horizon 2020 research and innovation programme under grant agreement No. 732328 (FashionBrain) and by the German Federal Ministry of Education and Research (BMBF) under grant agreement No.
01UG1735BX (NOHATE) and No. 01MD19003B (PLASS).
\end{acks}

\bibliographystyle{ACM-Reference-Format}
\bibliography{cikm2019}

\end{document}